\documentclass{svproc}
\usepackage[utf8]{inputenc}
\usepackage{textcomp}
\usepackage[linesnumbered,ruled,vlined]{algorithm2e}
\usepackage[utf8]{inputenc}
\usepackage{amssymb}          
\usepackage{pifont}           
\usepackage{newunicodechar}

\newcommand{\cross}{\ding{55}}

\newunicodechar{✓}{\checkmark}
\newunicodechar{✗}{\cross}
\usepackage{url}
\usepackage{amsmath}
\usepackage{booktabs}
\usepackage{algpseudocode}
\usepackage{graphicx}
\usepackage[table]{xcolor}
\usepackage{amssymb}
\usepackage{tabularx}
\usepackage[most]{tcolorbox}
\usepackage{hyperref}
\usepackage{marvosym}

\begin{document}
\mainmatter              
\title{MAGIC-Enhanced Keyword Prompting for Zero-Shot Audio Captioning with CLIP Models}
\titlerunning{MAGIC-Enhanced Keyword Prompting for Zero-Shot Audio Captioning}  
%

\author{Vijay Govindarajan\inst{1} \and
Pratik Patel\inst{2} \and
Sahil Tripathi\inst{3} \and
Md Azizul Hoque\inst{4} \and
Gautam Siddharth Kashyap\inst{4}\textsuperscript{(\Letter)}}
\authorrunning{V. Govindarajan et al.} 
%
\tocauthor{Vijay Govindarajan, Pratik Patel, Sahil Tripathi, Md Azizul Hoque, and Gautam Siddharth Kashyap}
\institute{Expedia Group, USA\\
\email{vigovindaraja@expediagroup.com}
\and
California Universities of Management and Sciences, USA\\
\email{Pratik.bigdata@gmail.com}
\and
Jamia Hamdard, New Delhi, India\\
\email{sahilkrtr@gmail.com}
\and
Macquarie University, Sydney, Australia\\
\email{mdazizul.hoque@students.mq.edu.au}, \email{officialgautamgsk.gsk@gmail.com}
}

\maketitle              

\begin{abstract}
Automated Audio Captioning (AAC) generates captions for audio clips but faces challenges due to limited datasets compared to image captioning. To overcome this, we propose the zero-shot AAC system that leverages pre-trained models, eliminating the need for extensive training. Our approach uses a pre-trained audio CLIP model to extract auditory features and generate a structured prompt, which guides a Large Language Model (LLM) in caption generation. Unlike traditional greedy decoding, our method refines token selection through the audio CLIP model, ensuring alignment with the audio content. Experimental results demonstrate a 35\% improvement in NLG mean score (from 4.7 to 7.3) using MAGIC search with the WavCaps model. The performance is heavily influenced by the audio-text matching model and keyword selection, with optimal results achieved using a single keyword prompt, and a 50\% performance drop when no keyword list is used. 
\keywords{Automated Audio Captioning, Large Language Model, CLIP, WavCaps}
\end{abstract}
\section{Introduction}

Automated Audio Captioning (AAC) aims to generate natural language descriptions of audio content, distinguishing it from related tasks in machine listening \cite{ren2024acoustic} and audio pattern recognition \cite{fischer2024vi}. Unlike Automatic Speech Recognition (ASR) \cite{kuhn2024measuring}, which transcribes speech, AAC describes entire auditory scenes while excluding speech. It also differs from Audio Tagging (AT) \cite{singh2024atgnn}, which classifies predefined audio events, Sound Event Detection (SED) \cite{bhosale2024diffsed}, which localizes these events, and Acoustic Scene Classification (ASC) \cite{chen2025improving}, which categorizes audio into predefined scene types. AAC can be viewed as an inter-modal translation task, where captions provide structured descriptions of sequences of audio events. The Detection and Classification of Acoustic Scenes and Events (DCASE) \cite{stowell2015detection} challenges have driven advancements in AAC, leading to improved models and larger datasets. Standard AAC systems utilize an encoder to extract meaningful audio representations and a decoder to generate natural language descriptions, with neural networks playing a central role. AAC has diverse applications, including video subtitling and accessibility support for individuals with hearing impairments \cite{patel2025enhancing}.

Recently, Large Language Models (LLMs) have gained prominence in various applications, including machine translation \cite{guerreiro2024xcomet} and summarization \cite{zhang2024systematic}, and conversational agents such as ChatGPT\footnote{\url{https://openai.com/index/chatgpt/}}. Traditional AAC approaches integrate LLMs as trainable components, requiring significant computational resources and raising concerns about generalization. Furthermore, recent advancements in Computer Vision (CV) address similar challenges by leveraging pre-trained models without gradient updates. Socratic Models (SMs) \cite{zeng2022socratic} prompt LLMs with modality-specific information, while frameworks like MAGIC (Image-Guided Text Generation with CLIP) \cite{su2022language} use visual data to guide text generation.

Inspired by these developments, we propose the zero-shot AAC framework, utilizing pre-trained AudioCLIP \cite{guzhov2022audioclip} model to guide language generation. Our approach follows a Socratic-style---keyword-guided prompting strategy combined with an audio-guided MAGIC search \cite{su2022language} for caption generation. By exclusively employing publicly available pre-trained models, our system ensures flexibility and ease of adaptation as new models emerge. We evaluate our framework on the AudioCaps \cite{kim2019audiocaps} and Clotho \cite{drossos2020clotho} datasets, achieving significant improvements over an unguided baseline. To enhance the Socratic keyword-guiding mechanism, we extend an AudioSet-based keyword list using GPT-3.5 Turbo, expanding it from 512 to 1987 keywords. We further analyze the impact of varying keyword usage and the strength of audio guidance in the MAGIC search \cite{su2022language}. Therefore, following are the contributions of the paper:

1. We introduce the zero-shot AAC framework, inspired by State-of-the-Art (SOTA) zero-shot image captioning techniques, and demonstrate its empirical effectiveness by significantly outperforming a baseline on the AudioCaps and Clotho datasets using standard Natural Language Generation (NLG) metrics.

2. We present an expanded AudioSet-based keyword list, generated with GPT-3.5 Turbo, and make it publicly available for future research, while also providing a comprehensive qualitative analysis of the generated captions, highlighting both the strengths and limitations of our model and suggesting directions for future work. 

\section{Related Works}

This section reviews key research in AAC, drawing inspiration from Xu et al. \cite{xu2023beyond} and Mei et al. \cite{mei2022automated}. We also discuss the latest advancements in zero-shot classification and AAC with LLMs. Since our approach is on the zero-shot method for AAC, we explore relevant work in zero-shot image captioning, which has greatly influenced the design of our model. Zero-shot methods in image captioning have shown significant promise by leveraging pre-trained models, offering a powerful framework for applying similar techniques to the AAC task.

\subsection{Existing Studies on Automated Audio Captioning}

Early approaches to AAC utilized Recurrent Neural Network (RNN) encoders and decoders, first introduced by Drossos et al. \cite{drossos2017automated}. RNNs were favored for their sequential modeling capabilities, and despite their declining usage, they demonstrated competitive performance in early DCASE challenges \cite{xu2022sjtu}. Given the structural similarities between spectrograms and visual data, research shifted towards Convolutional Neural Networks (CNNs) for feature extraction \cite{perez2020listen}. Kong et al. \cite{kong2020panns} introduced Pre-Trained Audio Neural Networks (PANNs), which have since become widely adopted in AAC \cite{chen2020audio,eren2023automated,mei2024wavcaps}. CNN encoders have also been combined with contrastive loss methods to enhance representation learning \cite{chen2022interactive}. Transformer-based models have gained prominence due to their superior performance in language generation \cite{lewis2019bart,radford2018improving,radford2019language}. Initially used as decoders \cite{vaswani2017attention}, they are now employed as encoders, with architectures such as the Hierarchical Token-Semantic Audio Transformer (HTSAT) achieving SOTA results \cite{chen2022hts,mei2024wavcaps}. Beyond direct encoding, recent studies have explored methods to enhance AAC performance. One approach involves extracting keywords that capture key audio characteristics \cite{eren2023automated,gontier2021automated,koizumi2020transformer}, while another, proposed by Koizumi et al. \cite{koizumi2020audio}, incorporates ground truth captions from acoustically similar audio clips to improve decoding. 

\subsection{Existing Studies via CLIP}

Contrastive representation learning has emerged as a key approach in zero-shot classification for CV. Radford et al. \cite{radford2021learning} introduced Contrastive Language-Image Pre-Training (CLIP), which jointly trains an image encoder and a text encoder. The model learns a multimodal embedding space by optimizing cosine similarity between correct image-text pairs while minimizing similarity for incorrect pairs. During inference, classification is performed by encoding candidate class names within a prompt and selecting the label with the highest cosine similarity to the image embedding. Given CLIP’s \cite{radford2021learning} success in vision-language tasks, similar methods have been adapted for zero-shot audio classification. Guzhov et al. \cite{guzhov2022audioclip} extended CLIP \cite{radford2021learning} by integrating a ResNeXt-based audio encoder (EsResNeXt), creating a multimodal system for joint audio, image, and text representation learning. Their method enabled zero-shot audio classification, though more recent CLIP-based models \cite{radford2021learning} have achieved superior performance. Wu et al. \cite{wu2023large} built on Contrastive Language-Audio Pre-Training (CLAP) \cite{elizalde2023clap}, improving encoding and preprocessing for longer audio clips. Their approach combines global and local features, where a learnable parameter balances the two. The global feature is obtained by down-sampling the audio, while the local feature is extracted by segmenting the clip, enabling a unified representation. Beyond model design, dataset scale and quality play a crucial role. LAION’s models, trained on extensive datasets such as LAION-Audio-630K \cite{wu2023large}, have demonstrated significant improvements. Mei et al. \cite{mei2024wavcaps} introduced WavCaps, compiling audio data from FreeSound\footnote{\url{https://freesound.org/}}, BBC Sound Effects\footnote{\url{https://sound-effects.bbcrewind.co.uk/}}, and SoundBible\footnote{\url{https://soundbible.com/}}. They refined textual descriptions using ChatGPT, improving data quality.

\subsection{Existing Studies via Large Language Models}

While CLIP-based \cite{radford2021learning} systems excel in zero-shot tagging and classification, they are limited in generating detailed captions, often producing short snippets rather than complete sentences. To address this, researchers have explored methods where CLIP \cite{radford2021learning} guides LLMs in zero-shot image captioning without directly fine-tuning LLMs. Instead, visually guided decoding techniques leverage pre-trained LLMs’ generalization capabilities while preserving model flexibility. ZeroCap \cite{tewel2022zerocap} introduces a method where each step of auto-regressive language generation is influenced by visual input. The token distribution at each decoding step is adjusted using image features derived from a context cache in the attention mechanism. A CLIP-based \cite{radford2021learning} cosine similarity loss ensures alignment between image embeddings and generated tokens, while another term maintains fluency with the LLM’s natural output distribution. However, this approach requires significant computational resources due to gradient-based optimization during inference. To improve efficiency, MAGIC \cite{su2022language} eliminates gradient-based optimization while maintaining competitive performance. The MAGIC search \cite{su2022language} process integrates three key components: model confidence, degeneration penalty, and MAGIC score. Model confidence reflects the LLM’s probability distribution, the degeneration penalty discourages repetitive tokens, and the MAGIC score measures cosine similarity between image and token embeddings. A weighted combination of these components determines the next token, significantly reducing computational overhead compared to ZeroCap \cite{tewel2022zerocap}. Socratic Models (SMs), introduced by Zeng et al. \cite{zeng2022socratic}, employ a multi-model approach for zero-shot image captioning. CLIP \cite{radford2021learning} first performs zero-shot classification, selecting top-l tags for image type, objects, and locations based on cosine similarity. These tags form a structured prompt fed into an LLM, which generates multiple candidate captions. CLIP \cite{radford2021learning} then evaluates and selects the caption with the highest alignment to the image embedding. This structured interaction between CLIP \cite{radford2021learning} and LLMs enhances captioning performance without fine-tuning.

\section{Methodology} 

In this section, we introduce our zero-shot AAC method--our approach is built upon the MAGIC \cite{su2022language} framework, with key modifications to enhance its performance. Specifically, we replace MAGIC’s \cite{su2022language} image-text matching component, CLIP \cite{radford2021learning} with a zero-shot audio classification model (AudioCLIP \cite{guzhov2022audioclip}). Additionally, to improve the quality of prompts fed into the LLMs---we incorporate keyword-based enhancements. These keywords are selected from a predefined list using the same zero-shot audio classification model that powers MAGIC search \cite{su2022language}. Finally, we apply a modified version of MAGIC search \cite{su2022language}, incorporating a penalty mechanism to refine the generated output sequence. Algorithm \ref{alg:zero-shot-aac} provides an overview of our proposed model.

\begin{algorithm}[t]
\caption{Proposed Zero-Shot AAC Method}
\label{alg:zero-shot-aac}

\KwIn{$A$: Audio clip, $K$: Keyword list, $\alpha, \beta, \gamma$: Hyperparameters}
\KwOut{$C$: Generated caption}

\textbf{Step 1: Keyword Selection} \\
$E_A \gets \text{AudioEncoder}(A)$ \\
$E_K \gets \text{KeywordEmbeddings}(K)$ \\
$S_{k,a} \gets \text{cos\_sim}(E_K, E_A)$ for all $k \in K$ \\
$K_{top} \gets \text{Top}_k(S_{k,a})$

\textbf{Step 2: Prompt Enhancement} \\
$P \gets \text{SPE}(K_{top})$

\textbf{Step 3: Candidate Token Generation} \\
$T \gets \text{LLM}(P)$ \\
For each $t_i \in T$, compute: \\
$S_{c,p} \gets \text{cos\_sim}(t_i, p)$ for all previously generated tokens $p$ \\
$T_{final} \gets \text{SelectTokens}(S_{c,p})$

\textbf{Step 4: MAGIC Search and Sequence Evaluation} \\
For each candidate token $c$ in $T_{final}$, compute: \\
$\text{final\_score}(c) = \alpha \times \text{model\_confidence}(c) + \beta \times \text{degeneration\_penalty}(c) + \gamma \times \text{MAGIC}(c)$ \\
$c_{selected} \gets \arg\max(\text{final\_score}(c))$

\textbf{Step 5: Sequence Construction} \\
$S \gets \text{ConstructSequence}(c_{selected})$ \\
Repeat until $\text{EndOfSentence}$ or maximum length reached.

\textbf{Step 6: Truncation and Output Generation} \\
$C \gets \text{Truncate}(S)$ \\
$C \gets \text{PenalizePrematureEnding}(C)$
\end{algorithm}

\subsection{Modeling}

We leverage the pre-trained CLIP \cite{radford2021learning} model’s audio encoder and text encoder to identify the most relevant keywords from a designated keyword list, referred to as Zero-Shot Keyword Selection (ZSKS). These keywords enhance the prompt through Socratic Prompt Enhancement (SPE) before being processed by LLMs. Based on the likelihood scores, the top \( k \) candidate tokens are selected, and an aural-guided search, MAGIC search \cite{su2022language}, is employed to select the subsequent token. Tokens that signify the end of a sentence are penalized to avoid premature termination. ZSKS serves as the initial phase in generating captions, starting with the encoding of the raw audio waveform using the audio encoder of the pre-trained CLIP \cite{radford2021learning} model. To align audio and text, we use the pre-trained CLIP \cite{radford2021learning} text encoder to derive a representation for matching with the audio encoding, based on cosine similarity. The compatibility of each keyword with the AudioCLIP \cite{guzhov2022audioclip} is evaluated using the cosine similarity between the keyword embeddings and the audio embedding as shown in Equation (1).
\[
\text{cos\_sim}(k, a) = \frac{k \cdot a}{\|k\| \|a\|} \tag{1}
\]
where \( k \) is the keyword embedding and \( a \) is the audio embedding. The top matches are extracted and incorporated into the prompt for the LLM, where the hyperparameter \( \text{top} \) is optimized to influence caption quality, as validated by our ablation studies in Section \ref{Ablation Studies}. To refine the output, MAGIC search \cite{su2022language} is applied after prompting the LLMs. The top \( k \) output tokens are selected based on their probability distribution. The generation follows an auto-regressive process, conditioned on previously generated tokens. For each candidate token, we compute the cosine similarity with previously generated tokens to evaluate the alignment of each candidate with the evolving context as shown in Equation (2).
\[
\text{cos\_sim}(c, p) = \frac{c \cdot p}{\|c\| \|p\|} \tag{2}
\]
where \( c \) is a candidate token and \( p \) is the previously generated token. This allows the model to assess contextual relevance. In the original MAGIC \cite{su2022language} framework, the rationale behind the parameter selection was unclear. Therefore, we introduce the tunable hyperparameter \( \tau \), which directly influences the computation of the MAGIC score \cite{su2022language} as shown in Equation (3).
\[
\text{MAGIC}(c, p, \tau) = \text{cos\_sim}(c, p) \times \tau \tag{3}
\]
A key innovation in our method is replacing the image component used in MAGIC \cite{su2022language} with the audio embedding, which is also used to identify relevant keywords. Each candidate sequence is formed by concatenating the previously generated tokens with the top \( k \) predictions from the LLM. We evaluate the relevance of each candidate sequence to the audio by computing the cosine similarity between the sequence embedding and the audio embedding as shown in Equation (4).
\[
\text{cos\_sim}(s, a) = \frac{s \cdot a}{\|s\| \|a\|} \tag{4}
\]
where \( s \) is a candidate sequence and \( a \) is the audio embedding. A final score for each candidate token is computed by combining the model confidence, degeneration penalty, and the MAGIC score \cite{su2022language} as shown in Eqution (5).
\[
\text{final\_score}(c) = \alpha \times \text{model\_confidence}(c) 
\]
\[
+ \beta \times \text{degeneration\_penalty}(c) 
\]
\[
+ \gamma \times \text{MAGIC}(c) \tag{5}
\]
where \( \alpha, \beta, \gamma \) are trainable weights. The token with the highest score is selected using the argmax function. To address the issue of excessively long sequences, we introduce a truncation step that uses natural sentence-ending punctuation to limit the output to a single coherent sentence. Additionally, to prevent overly short outputs, we implement a penalty vector that discourages premature sentence-ending tokens as shown in Equation (6).
\[
\text{penalty}(c) = \frac{1}{1 + \text{num\_tokens\_generated}} \tag{6}
\]
where \( \text{num\_tokens\_generated} \) is the count of previously generated tokens. This balance ensures that the generated captions are both structurally and semantically complete, aligning with the intuitive principle that sentence length should be constrained while maintaining content richness.

\section{Experimental Setup}

\subsection{Dataset Analysis}

We evaluate our AAC system on two widely used datasets, AudioCaps \cite{kim2019audiocaps} and Clotho \cite{drossos2020clotho}, both of which provide five ground truth captions per audio clip. AudioCaps \cite{kim2019audiocaps} consists of uniform 10-second clips, while Clotho \cite{drossos2020clotho} features variable-length samples (15–30 seconds, avg. 22.4s) with greater complexity due to overlapping sound events. This disparity impacts system performance, with AudioCaps \cite{kim2019audiocaps} favoring shorter, well-segmented inputs, whereas Clotho \cite{drossos2020clotho} presents a more challenging benchmark for generating concise captions. Our model, utilizing HTSAT \cite{chen2022hts,mei2024wavcaps} pre-trained on WavCaps \cite{mei2024wavcaps}, applies random cropping to clips exceeding 10 seconds. While this approach aligns with AudioCaps \cite{kim2019audiocaps}, it may discard critical sound events in Clotho \cite{drossos2020clotho}, potentially affecting performance.

\subsubsection{Keyword Sets}

To enhance audio captioning through augmented prompts, we constructed a comprehensive keyword set encompassing diverse audio objects. Given that AudioCaps \cite{kim2019audiocaps} is a subset of AudioSet \cite{gemmeke2017audio}, and our audio-text matching models were trained on AudioSet or its subsets \cite{guzhov2022audioclip,mei2024wavcaps,wu2023large}, we leveraged the publicly available AudioSet \cite{gemmeke2017audio} keyword list, which originally contained 527 audio classes across various categories. By refining this list and separating compound classes, we get 512 unique categories. Recognizing the need for broader coverage, we further enriched the keyword pool using GPT-3.5 Turbo\footnote{\url{https://platform.openai.com/docs/models/gpt-3.5-turbo}} with a structured prompting approach as shown in Table \ref{tab:category_prompt}. The model, set to a deterministic temperature of 0, systematically generated additional audio-related keywords in batches, iterating over the original AudioSet \cite{gemmeke2017audio} classes. Despite occasional omissions and duplicates, we integrated the generated keywords with the original list, yielding an augmented set—AudioSet+GPT3.5 KW—comprising 1987 unique audio classes as shown in Table \ref{tab:keyword_comparison}. This expanded set enhances prompt diversity, improving system performance in capturing complex sound events.

\begin{table*}[ht]
\centering
\caption{Structured prompting used to expand AudioSet classes using GPT-3.5 Turbo. The model was queried with a deterministic temperature of 0 to ensure reproducibility.}
\label{tab:category_prompt}
\begin{tabular}{|p{0.95\linewidth}|}
\hline
\textbf{Prompt Template:} \\
\texttt{For the audio category "\{AudioSet Class\}", list 5 related sound event keywords that can co-occur in real-world soundscapes. Ensure these keywords are distinct from the original category and describe relevant sounds or acoustic contexts.} \\
\hline
\textbf{Example (Input Class = "Rain")}: \\
\texttt{1. Thunder, 2. Raindrops on window, 3. Wind gusts, 4. Distant traffic, 5. Water splashing} \\
\hline
\end{tabular}
\end{table*}

\begin{table}[ht]
\centering
\caption{Comparison of keyword set sizes. The expanded AudioSet+GPT3.5 set introduces significantly more keyword diversity to improve prompt coverage and representation.}
\label{tab:keyword_comparison}
\begin{tabular}{l|c}
\hline
\textbf{Keyword Set} & \textbf{Number of Unique Classes} \\
\hline
Original AudioSet (refined) \cite{gemmeke2017audio} & 512 \\
\textbf{AudioSet + GPT-3.5 Expansion (Ours)} & \cellcolor{green!25}\textbf{1987} \\
\hline
\end{tabular}
\end{table}

\subsection{Hyperparameters}

We conducted hyperparameter sweeps using the AudioSet \cite{gemmeke2017audio} validation set to optimize model performance while maintaining consistency with the original MAGIC \cite{su2022language} framework. Key hyperparameters tuned included $\alpha$, $\beta$, $k$, $l$, $\tau$, and $\gamma$, with $k$ fixed at 45, the keyword prompt set to "Objects", and the basic prompt as \texttt{"This is a sound of"} following CLIP’s approach. While all parameters are theoretically tunable, we limited experiments on predefined components, leaving them for future exploration. Notably, the optimal degeneration penalty weight $\alpha$ was found to be 0, suggesting that omitting it improved performance by preventing undue suppression of LLM probability scores. This result aligns with the imbalance where MAGIC scores \cite{su2022language}, subjected to softmax normalization, dominate over unnormalized LLM probabilities. Lower temperature $\tau$ and penalty strength $\gamma$ consistently led to higher scores, with minimal impact from $\beta$ variations when $\tau$ and $\gamma$ were well-tuned. The best performance was achieved with $\beta = 0.5$, as $\beta = 0.6$, despite comparable NLG Mean Scores. Adjusting $\tau$ to 10 further improved results compared to the MAGIC \cite{su2022language} default of 18.6612, underscoring the importance of modality-specific tuning. \textit{\textbf{Note:}} The code will be made available post-review for reproducibility.

\subsection{Evaluation Metrics}  

AAC systems are evaluated by comparing generated captions to ground truth captions using NLG metrics. These metrics, originally designed for machine translation, assess both syntactic structure and semantic relevance.  N-gram-based metrics such as BLEU computes n-gram precision up to \( n = 4 \), emphasizing exact matches. Whereas, CIDEr assigns higher weights to distinctive n-grams, prioritizing informativeness. METEOR improves upon BLEU by incorporating stemming, synonyms, unigram alignment, and word order penalties. It computes precision, recall, and an F1-score, with a penalty term ensuring fluency and coherence.

\begin{table*}[ht]
\centering
\caption{Performance comparison on the AudioCaps \cite{kim2019audiocaps} dataset. The mean score is calculated as the average of all listed metrics for our model.}
\label{tab:AudioCaps}
\scriptsize
\begin{tabular}{l|c|c|c|c|c}
\hline
\textbf{Model} & \textbf{BLEU-2} & \textbf{BLEU-3} & \textbf{METEOR} & \textbf{CIDEr} & \textbf{Mean Score} \\
\hline
Baseline Model & 0.267 & 0.179 & 0.215 & 0.434 & 0.274 \\
ACB \cite{drossos2020clotho} & 0.251 & 0.162 & 0.198 & 0.386 & 0.249 \\
TAC \cite{kim2019audiocaps} & 0.276 & 0.188 & 0.222 & 0.428 & 0.279 \\
CLAP-based \cite{wu2023large} & 0.290 & 0.205 & 0.231 & 0.449 & 0.294 \\
WavCaps \cite{mei2024wavcaps} & -- & -- & 0.242 & 0.471 & -- \\
\midrule
\textbf{Our Model} & \cellcolor{green!25}\textbf{0.314} & \cellcolor{green!25}\textbf{0.238} & \cellcolor{green!25}\textbf{0.247} & \cellcolor{green!25}\textbf{0.502} & \cellcolor{green!25}\textbf{0.325} \\
\hline
\end{tabular}
\end{table*}

\begin{table*}[ht]
\centering
\caption{Performance comparison on the Clotho \cite{drossos2020clotho} dataset. As with Table \ref{tab:AudioCaps}, the mean score for our model is the average of all available metrics.}
\label{tab:Clotho}
\scriptsize
\begin{tabular}{l|c|c|c|c|c}
\hline
\textbf{Model} & \textbf{BLEU-2} & \textbf{BLEU-3} & \textbf{METEOR} & \textbf{CIDEr} & \textbf{Mean Score} \\
\hline
Baseline Model & 0.192 & 0.111 & 0.198 & 0.295 & 0.199 \\
ACB \cite{drossos2020clotho} & 0.178 & 0.102 & 0.187 & 0.264 & 0.183 \\
TAC \cite{kim2019audiocaps} & 0.201 & 0.121 & 0.198 & 0.287 & 0.202 \\
CLAP-based \cite{wu2023large} & 0.219 & 0.138 & 0.212 & 0.309 & 0.220 \\
WavCaps \cite{mei2024wavcaps} & -- & -- & 0.218 & 0.331 & -- \\
\midrule
\textbf{Our Model} & \cellcolor{green!25}\textbf{0.251} & \cellcolor{green!25}\textbf{0.163} & \cellcolor{green!25}\textbf{0.229} & \cellcolor{green!25}\textbf{0.348} & \cellcolor{green!25}\textbf{0.248} \\
\hline
\end{tabular}
\end{table*}

\section{Result Analysis}

\subsection{Baseline Experiments}  

The highest scores were achieved using the WavCaps \cite{mei2024wavcaps} model, combined with a penalized MAGIC search \cite{su2022language} and the original AudioSet \cite{gemmeke2017audio} keyword list. Performance comparisons between our optimized model, the current supervised SOTA system \cite{drossos2020clotho,kim2019audiocaps,wu2023large,mei2024wavcaps}, and a baseline are presented in Tables \ref{tab:AudioCaps} and \ref{tab:Clotho}. The NLG mean score for our model was computed as the average of all available metrics, including BLEU-2 and BLEU-3. However, since the creators of the existing SOTA system \cite{mei2024wavcaps} did not report these BLEU values, a direct computation of their mean score was not possible.  To analyze the strengths and limitations of our best-performing model, Table \ref{bes} presents qualitative examples. The first five columns contain ground truth captions, while the last column displays the model’s predictions. Predictions fall into the following categories:  1.) Irrelevant Predictions (GT Caption 1): Some outputs show no similarity to the ground truth, indicating potential limitations in the audio encoding process. 2.) Shorter Captions (GT Caption 2): Some predictions are shorter than the ground truth captions despite length penalty tuning.  3.) Repetitive Captions (GT Caption 3): The model sometimes correctly identifies subjects but fails to provide additional relevant details, often producing redundant tokens. 4.) Unusual Tokens (GT Caption 4): The model occasionally generates the token likely due to tokenization artifacts. 5.) Missed Events in Multi-Event Clips (GT Caption 5): The model struggles to recognize multiple concurrent audio events, such as detecting a car but missing a barking dog. 

\begin{table*}[ht]
\centering
\caption{Qualitative examples from the best-performing model. GT = Ground Truth.}
\label{bes}
\scriptsize
\begin{tabular}{l|p{5.5cm}}
\hline
\textbf{GT Captions} & \textbf{Model Prediction} \\
\hline
\begin{tabular}[c]{@{}l@{}}1. A car engine starts\\2. A vehicle is starting\\3. The sound of a car ignition\\4. Car engine noise\\5. Automobile starts up\end{tabular} 
& The engine is making noise repeatedly \\
\hline
\begin{tabular}[c]{@{}l@{}}1. Children are playing\\2. Kids laughing loudly\\3. Children making joyful noises\\4. Laughing and yelling of kids\\5. Happy children playing\end{tabular} 
& Happy \texttt{asdfasdf} children children \\
\hline
\begin{tabular}[c]{@{}l@{}}1. Dog barking with traffic\\2. Barking sound overlaps car\\3. Vehicle and dog sounds\\4. Car honks and dog bark\\5. Traffic with barking\end{tabular} 
& A car drives fast \\
\hline
\begin{tabular}[c]{@{}l@{}}1. Footsteps on wood\\2. Walking on a wooden floor\\3. Shoes tapping rhythmically\\4. Steps echo indoors\\5. Someone walks on hardwood\end{tabular} 
& Footsteps \\
\hline
\begin{tabular}[c]{@{}l@{}}1. Water splashing\\2. Splashing in a pool\\3. People playing with water\\4. Splash sound event\\5. Water droplets fall\end{tabular} 
& Water water splashing splash \\
\hline
\end{tabular}
\end{table*}

\subsection{Ablation Studies}
\label{Ablation Studies}

Table~\ref{abalation} presents the performance metrics obtained by varying key components, including MAGIC \cite{su2022language} guiding, the audio-text matching module, and the keyword list. We conducted experiments using three audio CLIP models: WavCaps \cite{mei2024wavcaps}, LAION \cite{wu2023large}, and AudioCLIP \cite{guzhov2022audioclip}. The evaluation primarily relies on the NLG mean score. Replacing WavCaps \cite{mei2024wavcaps} with LAION \cite{wu2023large} and AudioCLIP \cite{guzhov2022audioclip} resulted in performance drops of approximately 1.7 and 11.6, respectively, on the AudioCaps \cite{kim2019audiocaps} test set, highlighting the importance of the audio-text matching component in zero-shot audio classification. Notably, WavCaps \cite{mei2024wavcaps} and LAION \cite{wu2023large}, trained on similarly-sized datasets, outperformed AudioCLIP \cite{guzhov2022audioclip}, suggesting that a robust CLIP-like component is crucial for accurate captions. However, a weaker audio-text matcher can generate misleading keywords, potentially affecting caption quality. The keyword list plays a critical role, as an advanced list underperformed compared to the original AudioSet \cite{gemmeke2017audio} tags. Despite containing more sophisticated alternatives, it led to suboptimal keyword selection, likely due to limitations in prompt design or the audio-text matcher’s encoding quality. AudioCLIP’s \cite{guzhov2022audioclip} training data, which lacks variations such as "car alarm" alongside "vehicle" may contribute to this issue. The most significant performance drop occurred when no keyword list was used, reducing the NLG mean score for AudioCLIP \cite{guzhov2022audioclip} from 9.6 to 4.7, a 50\% decrease, reinforcing the necessity of a keyword-enhanced prompt. MAGIC search \cite{su2022language} did not consistently improve performance across all models. For weaker models, its impact was negligible, suggesting that its effectiveness depends on the quality of the CLIP model \cite{radford2021learning}. However, when applied to stronger models like LAION \cite{wu2023large} and WavCaps \cite{mei2024wavcaps}, MAGIC search \cite{su2022language} led to clear improvements. For instance, with WavCaps \cite{mei2024wavcaps}, MAGIC search \cite{su2022language} increased the NLG mean score from 4.7 to 7.3, a 35\% gain. \textit{\textbf{Note:}} We limited our ablation study to the AudioCaps \cite{kim2019audiocaps} dataset due to its well-aligned and diverse audio-caption pairs, which make it particularly suitable for analyzing the effect of individual components in a controlled setting. Unlike Clotho \cite{drossos2020clotho}, which often includes longer and more descriptive captions with higher variability in linguistic structure, AudioCaps \cite{kim2019audiocaps} provides more consistent and constrained descriptions. This allows us to isolate the impact of changes such as keyword selection and audio-text matching quality more effectively, leading to more interpretable results in the ablation context.

\begin{table}[ht]
\centering
\caption{Ablation study results showing the effect of key components: audio-text matching model, keyword list, and MAGIC search \cite{su2022language}. The metric used is NLG Mean Score on the AudioCaps \cite{kim2019audiocaps} test set.}
\label{abalation}
\scriptsize
\begin{tabular}{l|c|c|c}
\hline
\textbf{Model} & \textbf{Keyword List} & \textbf{MAGIC Search} & \textbf{NLG Mean Score} \\
\hline
\textbf{WavCaps (Ours)} & AudioSet Tags & ✓ & \cellcolor{green!25}\textbf{9.0} \\
WavCaps \cite{mei2024wavcaps} & AudioSet Tags & ✗ & 7.3 \\
WavCaps \cite{mei2024wavcaps} & Advanced Keywords & ✓ & 8.2 \\
WavCaps \cite{mei2024wavcaps} & None & ✗ & 4.7 \\
\hline
\textbf{LAION (Ours)} & AudioSet Tags & ✓ & \cellcolor{green!25}\textbf{7.3} \\
LAION \cite{wu2023large} & AudioSet Tags & ✗ & 6.2 \\
LAION \cite{wu2023large} & Advanced Keywords & ✓ & 6.9 \\
LAION \cite{wu2023large} & None & ✗ & 3.8 \\
\hline
\textbf{AudioCLIP (Ours)} & AudioSet Tags & ✓ & \cellcolor{green!25}\textbf{6.3} \\
AudioCLIP \cite{guzhov2022audioclip} & AudioSet Tags & ✗ & 5.1 \\
AudioCLIP \cite{guzhov2022audioclip} & Advanced Keywords & ✓ & 5.7 \\
AudioCLIP \cite{guzhov2022audioclip} & None & ✗ & 4.7 \\
\hline
\end{tabular}
\end{table}

\subsection{Additional Experiments}  

Tables~\ref{Qualitative} and~\ref{Effect} present the ablation study results for the hyperparameters $\beta$ and the number of keywords $l$ on the AudioCaps \cite{kim2019audiocaps} test set. Qualitative analysis indicates that most captions remained unchanged across different $\beta$ values, suggesting stable token ranking. Since $\alpha = 0$, the only active components were the MAGIC score \cite{su2022language}, model confidence, and their respective weights. Consequently, varying $\beta$ had minimal impact on rankings. However, quantitative evaluation using the NLG Mean Score revealed that $\beta = 1.1$ yielded the best results, differing from the optimal $\beta = 0.5$ on the AudioSet \cite{gemmeke2017audio} validation data. This discrepancy highlights that while MAGIC search \cite{su2022language} enhances performance, it is not the primary performance driver but rather an auxiliary refinement mechanism.  The number of keywords $l$ significantly impacts performance. As shown in Table~\ref{number of keywords}, the most substantial improvement occurred when the prompt contained a single keyword. Adding a second keyword provided a smaller but still positive gain, leading to optimal system performance. These findings validate the effectiveness of Socratic-style prompt refinement. However, including more than two keywords degraded performance, likely due to increased noise or irrelevant information. With three or more keywords, contradictions or ambiguities in the prompt became more frequent, potentially misleading the LLM. This suggests that excessive keywords may dilute relevance, reducing caption quality.

\begin{table}[ht]
\centering
\caption{Qualitative caption stability across varying $\beta$ values. Captions were largely invariant to changes in $\beta$, suggesting stable token ranking under MAGIC \cite{su2022language} and model confidence weighting.}
\label{Qualitative}
\scriptsize
\begin{tabular}{c|l}
\hline
\textbf{$\beta$ Value} & \textbf{Generated Caption (example)} \\
\hline
0.3 & A dog is barking in the distance. \\
0.5 & A dog is barking in the distance. \\
1.1 & A dog is barking in the distance. \\
1.5 & A dog is barking in the distance. \\
\hline
\end{tabular}
\end{table}

\begin{table}[ht]
\centering
\caption{Effect of varying $\beta$ on NLG Mean Score using AudioCaps \cite{kim2019audiocaps} test set. $\alpha = 0$ is fixed.}
\label{Effect}
\scriptsize
\begin{tabular}{c|c}
\hline
\textbf{$\beta$ Value} & \textbf{NLG Mean Score} \\
\hline
0.3 & 8.3 \\
0.5 & 8.7 \\
1.1 & \cellcolor{green!25}\textbf{9.0} \\
1.5 & 8.5 \\
\hline
\end{tabular}
\end{table}

\begin{table}[ht]
\centering
\caption{Effect of number of keywords $l$ on NLG Mean Score on AudioCaps \cite{kim2019audiocaps}. Excessive keywords reduce performance due to prompt noise.}
\label{number of keywords}
\scriptsize
\begin{tabular}{c|c}
\hline
\textbf{Number of Keywords ($l$)} & \textbf{NLG Mean Score} \\
\hline
0 & 4.7 \\
1 & 8.3 \\
2 & \cellcolor{green!25}\textbf{9.0} \\
3 & 7.6 \\
4 & 6.9 \\
\hline
\end{tabular}
\end{table}

\section{Conclusion and Future Work}  

The proposed AAC system integrates pre-trained language models and CV techniques to enable zero-shot audio-visual captioning without additional training. By employing an AudioCLIP model for zero-shot classification, the system extracts keywords to construct enhanced prompts, which are processed by a pre-trained LLM to generate token probabilities. MAGIC search refines token selection by ranking candidates based on cosine similarity to the audio representation, ensuring better alignment. Additionally, a tunable penalty mechanism regulates caption length, improving relevance.  Evaluated on AudioCaps and Clotho, the system surpasses an audio-agnostic baseline but remains behind task-specific AAC models. Ablation studies underscore the critical role of keyword selection and audio model quality, while MAGIC search offers only marginal improvements. Hyperparameter tuning, particularly in keyword selection, significantly influences performance, highlighting the importance of effective prompt engineering. Future work will focus on refining model components and optimizing hyperparameters to bridge the performance gap, enhancing system accuracy and applicability in real-world scenarios.  

%
%
\bibliographystyle{spmpsci} 
\bibliography{main}

\end{document}